\documentclass[runningheads]{llncs}
\usepackage[T1]{fontenc}
\usepackage{graphicx}
\usepackage{subcaption}
\usepackage{booktabs}
\usepackage{tabularx}
\usepackage{xcolor}
\usepackage{caption}
\begin{document}
\title{GA-VisAgent: A Multi-Agent application for code generation and visualization in interactive learning}
%
%
\author{Wang Jian\inst{1} \and
Zhou Jianbo\inst{1} \and
Xiong Yuhao\inst{1} \and
Liu Zhenxia\inst{2} \and
Luo Wen\inst{3,4} \and
Yuan LinWang\inst{3,4,5} \and
Yu ZhaoYuan\inst{3,4,5}\thanks{Corresponding author}}

\authorrunning{W. Jian et al.}
%
\institute{
School of Geography, Nanjing Normal University, Nanjing 210023, Jiangsu, China
\and  
School of Environment, Nanjing Normal University, Nanjing 210023, Jiangsu, China
\and
Key Laboratory of Virtual Geographic Environment, Ministry of Education, Nanjing Normal University, Nanjing 210023, Jiangsu, China
\and  
State Key Laboratory Cultivation Base of Geographical Environment Evolution, Nanjing 210023, Jiangsu, China
\and  
Jiangsu Center for Collaborative Innovation in Geographical Information Resource Development and Application, Nanjing 210023, Jiangsu, China
\email{yuzhaoyuan@njnu.edu.cn}}

\maketitle              
\begin{abstract}
\textcolor{blue}{Geometric Algebra (GA) presents challenges to learners due to its highly abstract mathematical structure and complex operational rules, as translating algebraic manipulations into concrete geometric interpretations is a non-intuitive process when developing related code.} Currently, some existing GA software packages rely on manually written scripts for code generation and visualization, but their high learning curve hinders widespread adoption. Meanwhile, methods based on Large Language Models (LLMs) often produce logical errors when generating specific GA scripts, such as GAALOPScript, resulting in generally low accuracy. To address these issues, this study proposes GA-VisAgent—a multi-agent interactive learning application for GA code generation and visualization—building upon a Geometric algebra large language model(GAGPT). Integrating task planning mechanisms with ReAct reasoning strategies, GA-VisAgent can decompose complex operations into five standardized subtasks, including core operations like geometric products, rotations, and reflections. It supports natural language and mathematical formulas as input to automatically generate executable code, accompanied by interactive visualizations to aid user comprehension. Experimental results show that GA-VisAgent achieved a 90\% code generation success rate across 40 typical Conformal GA tasks, representing a 70\% improvement over GPT-4o. This application introduces an extensible new paradigm for teaching GA and developing visualization tools for related mathematical concepts. The online service for this project will be available at http://gagis.cn/gacrac.

\end{abstract}
\section{Introduction}
Geometric algebra (GA) serves as a unified language in mathematics, physics, and engineering\cite{fontijne2007geometric,hestenes2012new,angelis2024geometric}. GA is widely used in various fields, such as computer graphics\cite{hadfield2021exploring}, neural networks\cite{li2022ga,thiruvengadam2020time}, pattern recognition\cite{labunets2004clifford},  and signal and image processing\cite{hitzer2022foundations,orouji2019hardware}. GA presents challenges to learners due to its highly abstract mathematical structure and complex operational rules, making the process of translating algebraic manipulations into concrete geometric interpretations a non-intuitive one when developing related code.

Interactive dynamic code generation and visualization tools can enhance learners' engagement and efficiency during the learning process. However, the interactive visualization tools provided in current GA textbooks present certain barriers to use and require learners to acquire additional knowledge. For instance, tools like Geometric algebra algorithms optimizer(GAALOP)\cite{hildenbrand2010gaalop}, CLUCalc\cite{perwass2009learning}, and \textcolor{blue}{Ganja.js\cite{ganja.js}} demand learners to write specialized scripting languages (e.g., GAALOPScript\cite{alves2020efficient}) to generate code and visualizations, necessitating separate training in these domain-specific languages and thereby increasing the entry difficulty for beginners. The emergence of large language models (LLMs), capable of interpreting learners' natural language inputs and generating corresponding knowledge, holds the potential to assist novices in learning GA. However, current LLMs primarily target general domains and lack optimization for GA, frequently producing errors when outputting scripts for GA software packages like GAALOPScript, CLUScript, or Ganja. Additionally, comprehending GA's intricate computational processes remains challenging. Furthermore, due to the absence of GA knowledge rules for constraint, the accuracy and reliability of code generated by Generative Pre-trained Transformer(GPT) models cannot be guaranteed. These limitations become particularly pronounced when handling typical complex GA tasks, where generated code often contains various syntax and logical errors.

LLMs are systems capable of generating text and answering questions in natural language\cite{liang2024survey}. Recent advancements in LLM agents have significantly enhanced their capabilities in handling complex tasks. LLMs agent is defined as a framework composed of three components: a brain, perception, and action\cite{wang2024survey}. Architectures can be classified as single-agent or multi-agent systems, depending on the number of agents involved\cite{heik2024study}. The ReAct framework\cite{yao2023react} exemplifies this by enabling agents to first reason about tasks and then execute actions, outperforming traditional zero-shot prompting methods. Building upon ReAct, the agent framework enhances LLM performance in vertical domains by decomposing complex tasks and integrating tool invocation capabilities. The fusion of LLMs with conformal geometric algebra (CGA) can automate precise 3D scene editing tasks, such as object repositioning, which traditionally required labor-intensive manual workflows and domain-specific expertise\cite{angelis2024geometric}; Geometric algebra large language model(GAGPT) enhances LLMs' computational capabilities and knowledge-based question-answering proficiency in GA by integrating LLMs with GA knowledge\cite{wang2023large,Wang2025}.

Based on the multi-agent collaborative framework, this study significantly improves the code generation capability of LLMs in the field of GA by decomposing complex GA formulas into five categories of subtasks and building a dedicated function library and Application Programming Interface(API) calling mechanism. The application adopts a two-layer architecture of planner agent and work agent, dynamically analyzes user needs through the ReAct mechanism, and converts high-dimensional operations into structured task flows. This study uses GAGPT as the base model and guides LLMs to generate precise code that conforms to GAALOPScript syntax for 30 dedicated functions of each subtask, ultimately achieving a 90\% success rate in conformal space tasks.

\textcolor{blue}{This paper is structured as follows: Section 2 introduces the core concepts and multi-agent methodology of the Multi-Agent Code Generation and Visualization Application for GA; Section 3 details its architecture, including agent coordination and subtask taxonomy; Section 4 covers implementation via ReAct scheduling, libraries, and the code pipeline; Section 5 validates efficacy on 40 conformal space tasks; and Section 6 concludes with plans for integrating GA software packages.}

\section{Idea}
The core challenge in teaching practices for the application of GA lies in constructing a cognitive bridge between abstract mathematical theories and engineering implementations. This paper focuses on a triadic "theory-code-geometric visualization" learning loop model, where learners input natural language descriptions and GA mathematical formulas, and the system synchronously outputs executable code and interactive dynamic visualization results. However, existing methods face three bottlenecks: first, traditional manual coding requires learners to simultaneously master GA algebraic rules and syntactic details of domain-specific scripting languages (e.g., GAALOPScript/CLUScript), leading to cognitive overload; second, code generated by generic LLMs often contains logical errors or formatting deviations due to insufficient domain-specific syntactic constraints; third, static visualization schemes struggle to depict intermediate-state evolution processes in high-dimensional geometric transformations, hindering in-depth verification and understanding of operators' geometric implications.

This study proposes the GA-VisAgent framework to address these challenges, with its innovation primarily reflected in dual collaborative mechanisms: First, a dual-agent hierarchical architecture decouples complex GA tasks into atomic operational units. A Planner agent performs structured decomposition based on formula semantics, while a Worker agent generates syntactically compliant GAALOPScript/CLUScript code through domain-customized Agents and ReAct mechanisms, effectively mitigating semantic deviations in end-to-end LLMs generation. Second, a dynamic visualization embedding paradigm automatically injects script language-specific tracing statements during code generation, invoking the GAALOP Web API to synchronously generate executable code and interactive dynamic visualization results.

\section{GA-VisAgent Application Architecture}
\subsection{Application Architecture}
\begin{figure}[htbp]
    \vspace{-20pt}
    \centering
    \includegraphics[width=1.0\textwidth]{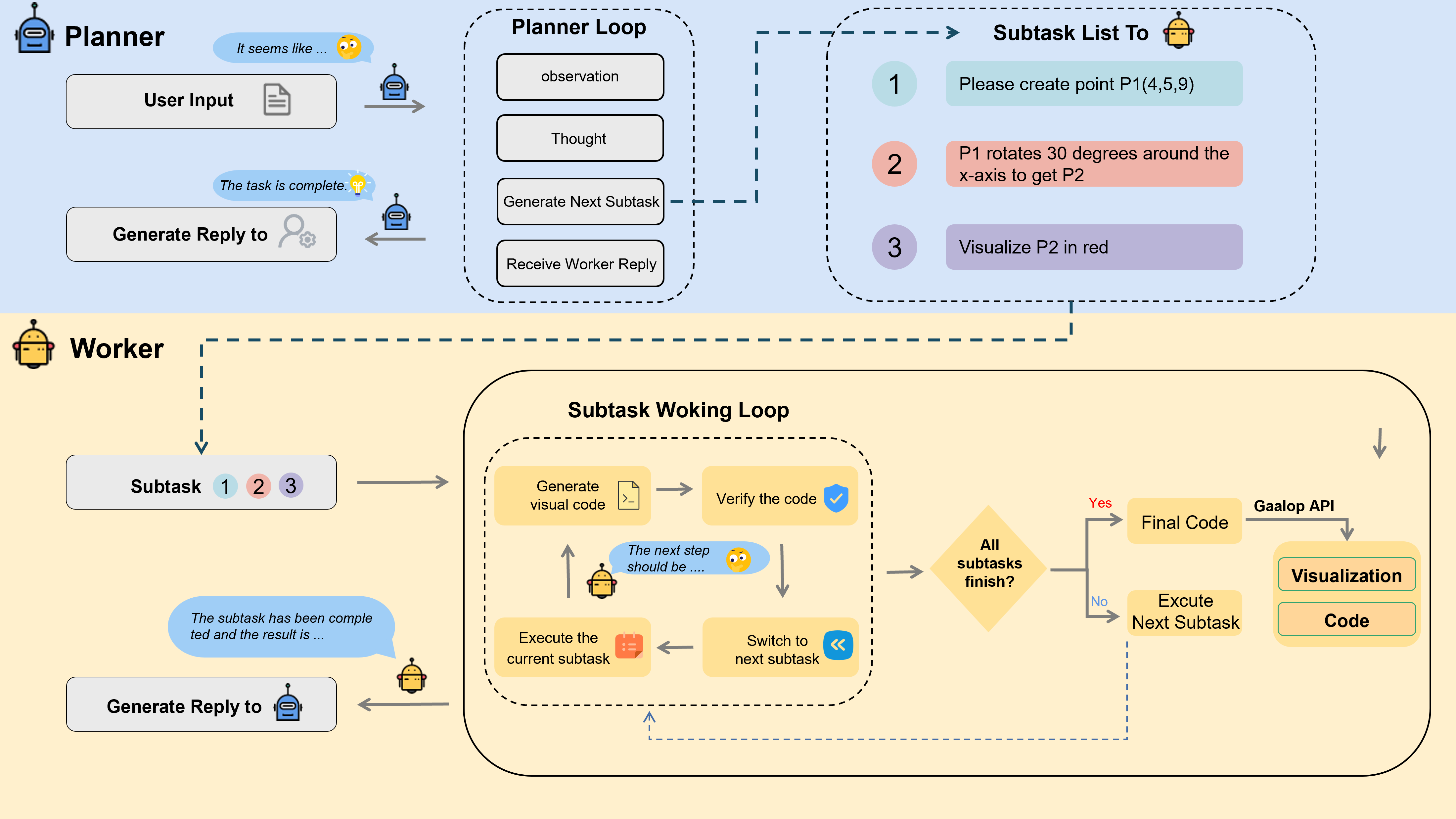}
    \caption{GA-VisAgent Application Architecture}
    \label{fig:1}
    \vspace{-20pt}
\end{figure}

GA-VisAgent converts natural language instructions and GA formulas into interactive code and dynamic visualizations.  It progressively processes these inputs to solve GA problems, generating the corresponding code and visual outputs. Figure~\ref{fig:1} illustrates the application architecture of GA-VisAgent. The system adopts a dual-agent cooperative framework: the planner agent acts as the decision-making center, receiving user instructions and decomposing them into manageable sub-tasks. The worker agent is responsible for carrying out these sub-tasks, generating the corresponding code, and producing interactive visualization results. Efficient communication between the two agents is achieved through an internal API, enabling seamless collaboration in automated task processing.

As the core hub of code generation, the Planner agent is responsible for parsing the GA formulas input by users. According to the mathematical structure of the GA formula and the user's visualization requirements, it is decomposed into multiple ordered GA sub-tasks. Each subtask corresponds to each component of the formula one-to-one, and the execution sequence between subtasks is strict.

The Worker Agent is the execution module for code generation and interactive visualization, utilizing a structured process to convert natural language and formulas into code and visualization results. This module first analyzes the subtasks output by the planner agent, identifying the type of geometric operations and visualization requirements within each subtask; then, it generates standardized code and visualization results based on GAALOPScript syntax and CLUCalc syntax. The modified code includes an optimization code section, a parameter assignment section, and a visualization section; finally, after passing a syntax check, it outputs executable code in a standardized format. 

After all subtasks are completed, GA-VisAgent will further call the GAALOP API based on spatial and other parameter information, converting the generated GAALOPScript script into GA code and visualization results, thereby completing the entire generation process.

\subsection{Complex task decomposition}
Aiming at the code generation and visualization requirements of complex GA formulas, this study uses task decomposition to improve execution efficiency and code accuracy. Since complex GA formulas often include multiple interrelated geometric operation steps, the overall processing can easily lead to a surge in computing complexity and it is difficult to ensure accuracy.

This paper adopts a task decomposition strategy based on geometric operation types (such as geometric object creation, basic algebraic operations, and geometric element manipulation, as detailed in Table 1), guided by three core design principles: First, decomposing fundamental components of geometric algebraic operations into atomic-level subtasks significantly enhances system execution efficiency and reliability. Second, this classification establishes a modular debugging framework that enables both rapid localization of anomalous subtasks for precise troubleshooting and confinement of modification scopes to individual subtasks, effectively mitigating systemic risks. Third, by systematically categorizing GA tasks into five distinct types—ranging from basic geometric element generation to complex geometric transformation combinations—a hierarchical task decomposition framework is established, providing structured support for subsequent code generation. This multi-granularity decomposition mechanism ensures the accuracy of executable code generation and interactive dynamic visualization results in GA-VisAgent through independent subtask verification protocols.

\begin{table}[htbp]
\vspace{-30pt}
\centering
\caption{Subtask Category Classification Framework}
\label{tab:subtask_classification2}
\begin{tabular}{lll}
\toprule
Action Category & Action Name & Mathematical Expression \\
\midrule
Fundamental Algebraic Operations & Geometric Product & $AB = A \cdot B + A \wedge B$ \\
& \dots & \dots \\
Geometric Element Operations & Reverse & $\widetilde{A} = (-1)^{k(k-1)/2}A$ \\
& \dots & \dots \\
Geometric Transformation & Projection & $\mathrm{proj}_B(A) = (A \cdot B)B^{-1}$ \\
& \dots & \dots \\
Numerical Operations & Normalization & $A_{\mathrm{norm}} = \frac{A}{\|A\|}$ \\
& \dots & \dots \\
Geometry Object Creation & Point Creation & $\mathbf{P} = x\mathbf{e}_1 + y\mathbf{e}_2 + z\mathbf{e}_3$ \\
& \dots & \dots \\
\bottomrule
\end{tabular}
\vspace{-20pt}
\end{table}

\begin{figure}[htbp]
\vspace{-20pt}
    \centering
    \captionsetup[subfigure]{
        labelfont=bf, 
        textfont=small, 
        justification=centering
    }
    
    \begin{subfigure}[t]{0.45\textwidth}
        \includegraphics[width=\linewidth, height=5cm]{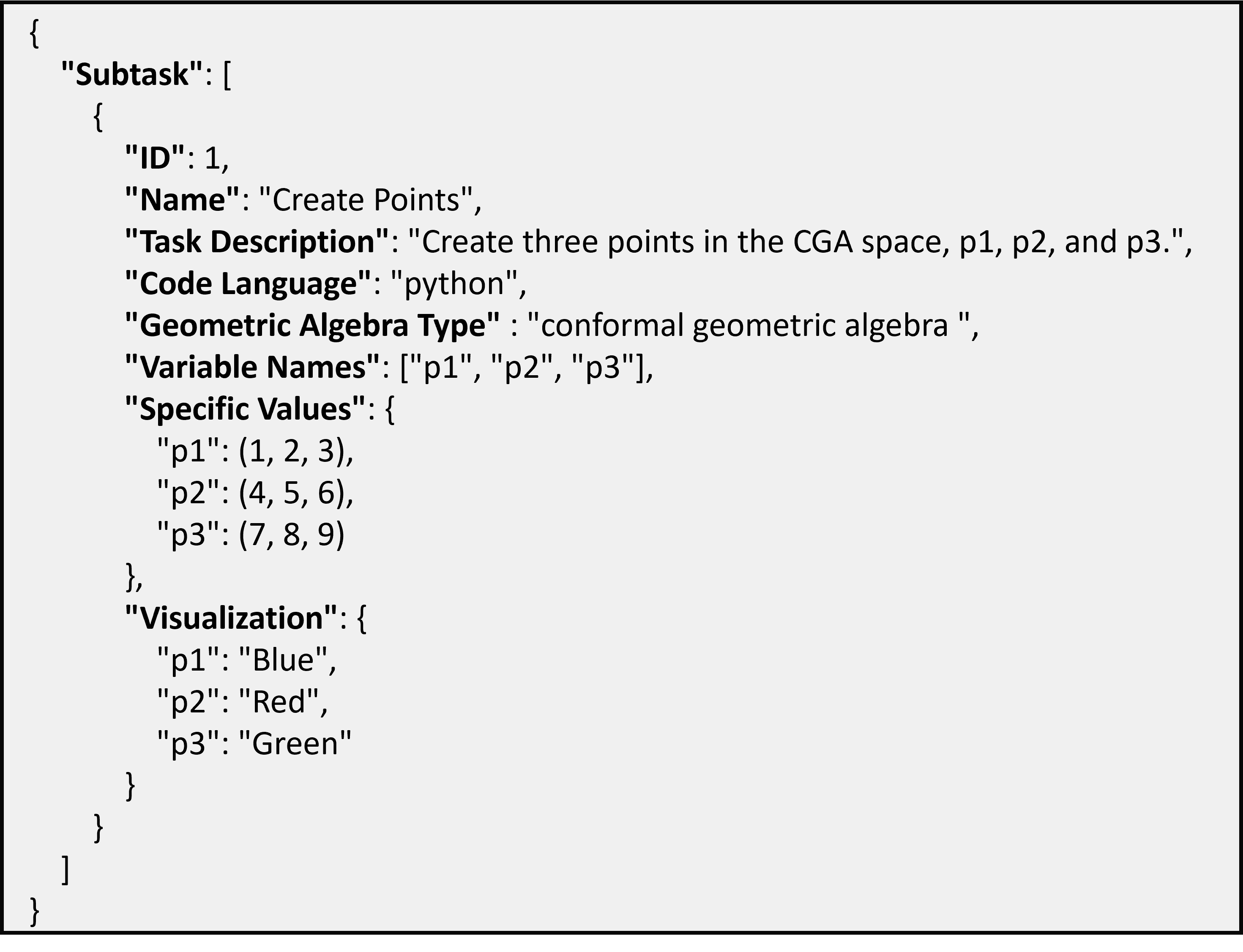}
        \caption{Subtask output format}
        \label{fig:doc-design}
    \end{subfigure}
    \hspace{0.5cm}
    \begin{subfigure}[t]{0.45\textwidth}
        \includegraphics[width=\linewidth, height=5cm]{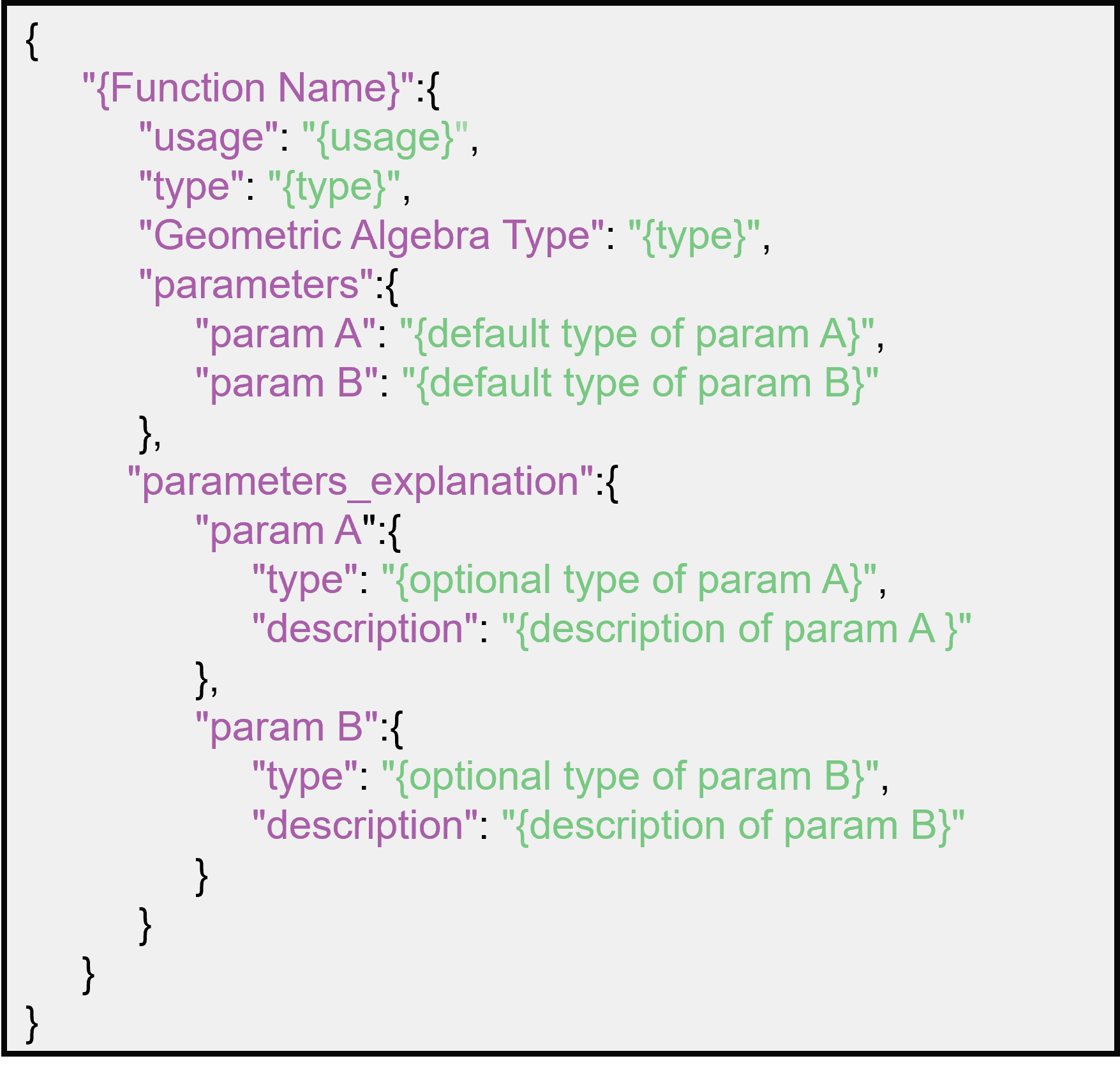}
        \caption{Function documentation design}
        \label{fig:output-format}
    \end{subfigure}
    
    \vspace{0.4em}
    \captionsetup{font=small, skip=6pt}
    \caption{Document design and output format specifications}
    \label{fig:both}
    \vspace{-35pt}
\end{figure}

To ensure that the Agent can read and execute sub-task information efficiently and accurately, this study sets up the sub-task output format. This format uses a structured definition that details key information such as task ID, task name, task description, variable names, code language, GA type, specific values, and visualization settings. The task ID is used as a unique identifier, which is convenient for the system to accurately locate and manage tasks, and quickly trace the source of problems; Task description helps the Agent to quickly understand the core requirements of tasks and avoid execution deviations; Variable names and specific values provide accurate data support to prevent variable assignment errors; Visual settings ensure that the results meet user expectations. Through this structured definition, the Agent can resolve and execute tasks more efficiently, and the uncertainty in task resolution and execution. The input format definition is illustrated in Figure \ref{fig:both}(\subref{fig:doc-design}).

\subsection{Function call}
\textcolor{blue}{This study leverages subtask decomposition and function calls to enable LLMs to perform GA formula conversion tasks. Function calls enhance LLM interaction with external tools, allowing models to invoke predefined functions during text generation. This mechanism extends LLM capabilities beyond natural language generation to interact with programs, databases, and APIs for complex tasks like computation, real-time data access, and specialized operations.}

\textcolor{blue}{While online LLMs like OpenAI's GPT-4o often rely on generating and executing code in virtual environments to accomplish tasks, our experiments show it frequently fails at accurate GA formula conversion. To address this, we design GA-VisAgent, which adopts a function calling mechanism. Crucially, the functions are specifically designed to guide LLMs toward precise invocation for improved conversion accuracy.}

\textcolor{blue}{Within GA-VisAgent, we developed a dedicated set of functions, along with corresponding libraries and APIs, specifically for GA formula processing. Detailed API documentation clearly outlines usage and parameter rules, enabling GA-VisAgent to select the appropriate API for each decomposed subtask. These functions equip the LLM with the necessary capabilities to effectively process each subtask. The structured function definitions are shown in Figure \ref{fig:both}(\subref{fig:output-format}).}


\section{GA-VisAgent Technical Architecture}
\begin{figure}
    \centering
    \vspace{-10pt}
    \includegraphics[width=0.6\linewidth]{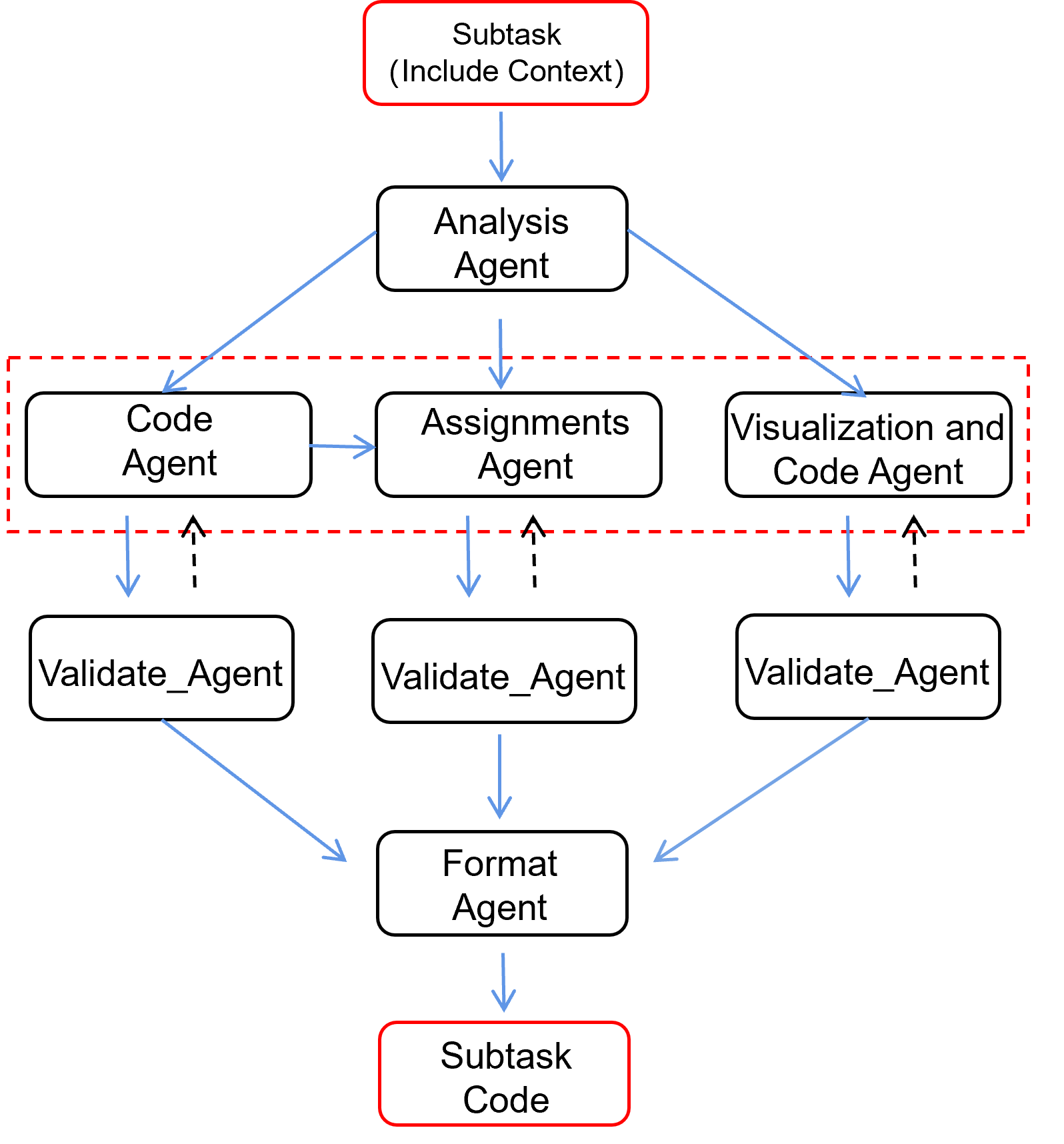} 
    \caption{Document Design of Function Library}
    \label{fig:4}
    \vspace{-20pt}
\end{figure}

\textcolor{blue}{Within the domain of GA code and visualization generation, GA-VisAgent achieves its core functionality through the dynamic generation of GAALOPScript code. The system's technical essence lies in constructing an efficient and precise code generation mechanism. This study innovatively employs a multi-agent collaborative architecture with function invocation mechanisms, demonstrating significant advantages over traditional single-agent systems. While single-agent systems often suffer from task overload, leading to chaotic responses and reduced efficiency when handling complex GA problems, the proposed multi-agent system utilizes a distributed task decomposition mechanism. This mechanism breaks down workflows into chains of subtasks executed collaboratively by specialized agents (detailed in Table~\ref{tab:subtask_classification1}). Each agent focuses on specific elementary tasks, enhancing overall efficiency through parallel processing and dynamic coordination (illustrated in Figure ~\ref{fig:4}'s function call process). Each subtask and its context are first processed by the Analysis Agent. This agent extracts key elements—including geometric objects, operation types, anonymous object naming rules, and related formulas—through structured information extraction. This process provides standardized data inputs for subsequent visualization and code generation, enabling all agents to parse and process tasks more efficiently.}

\begin{table}[htbp]
\centering
\caption{Subtask Category Classification}
\label{tab:subtask_classification1}
\begin{tabularx}{\textwidth}{l>{\raggedright\arraybackslash}X}
\toprule
\textbf{Agent Role} & \textbf{Core Responsibilities} \\
\midrule
Analysis Agent      & Analyze user input and extract key information in a structured manner \\
\addlinespace[2pt]
Code Agent          & Optimize core components of GAALOPScript code generation \\
\addlinespace[2pt]
Assignment Agent    & Extract user-input numerical values and assign them to parameters in optimized code \\
\addlinespace[2pt]
Visualization Agent & Visualize geometric objects per user requirements and generate tailored programming code \\
\addlinespace[2pt]
Validate Agent    & Verify GAALOPScript code compliance with syntax specifications \\
\addlinespace[2pt]
Format Agent        & Output results in specified formats \\
\bottomrule
\end{tabularx}
\vspace*{-5ex}  
\end{table}

\subsection{Core Agent Module}
\textcolor{blue}{After receiving and processing user input into subtask input, the Analysis Agent forwards its output results to three core agents: the Code Agent, Assignment Agent, and Visualization Agent. These agents collaboratively generate the three major components of the final GAALOPScript code: the optimization code, parameter assignment section, and visualization section.}

\textcolor{blue}{Code Agent: Based on GAALOP syntax specifications, this agent invokes corresponding functions to convert the input mathematical formulas into the optimized computational component of GAALOPScript code. It achieves precise mapping of formula operators to their script counterparts.}

\textcolor{blue}{Assignment Agent: This agent receives outputs from both the Analysis Agent and the Code Agent. Its primary function is to identify values and variable names within the Code Agent's output and the user input, subsequently generating the necessary assignment statements.}

\textcolor{blue}{Visualization and Code Agent: According to user requirements, this agent is responsible for visualizing geometric objects within the code and generating the corresponding visualization code. This process enables the transformation of abstract mathematical expressions into visual representations. Together with the Code Agent, it facilitates a complete pipeline from abstract concepts to practical computing code, allowing users to intuitively grasp geometric structures while achieving efficient computational implementation.}

\subsection{Code Validation and Formatting}
The generated code is then passed to the Validate Agent. The Agent is responsible for verifying that the code generated by the three core Agents conforms to the GAALOPScript syntax specification. If the code does not conform to the specification, Validate Agent will return it to the corresponding agent to regenerate the corresponding code, ensuring that the final output code. This step is a critical part of the code generation process because it ensures the accuracy and reliability of the code. After verification, the code is finally entered into the Format Agent. This agent is responsible for formatting the final result, ensuring that the code is output in a format that complies with the subsequent GAALOP API.

\section{Experiments and evaluations}

\subsection{Experimental Setup}
\textcolor{blue}{To evaluate the capabilities of GPT-4o and GA-VisAgent in processing various GA formulas and performing related operations, this study constructs a test dataset containing 40 formulas within the CGA space. The dataset covers core operation types such as GA construction and rigid motion transformation (partial examples are shown in Table~\ref{tab:Test}). The evaluation criterion is whether the generated GAALOPScript code (default output language: Python) can execute correctly. For GA-VisAgent, inputs consist of the "Task Description" content and Formulas directly, While GPT-4o requires the specific prompt template table as shown in Table~\ref{tab:Prompt template}).}

\begin{table}[ht]
\vspace{-20pt}
\caption{Test Dataset for GA Code Generation and Visualization}\label{tab:test-dataset}
\centering
\label{tab:Test}
\setlength{\tabcolsep}{5pt}
\renewcommand{\arraystretch}{1.3}
\begin{tabular}{>{\raggedright}p{52mm}>{\centering\arraybackslash}p{60mm}}
\toprule
\textbf{Task Description} & \textbf{GA Formula} \\
\midrule
Create point $p_1(4,5,6)$ (color: blue) 
    & $P = xe_1 + ye_2 + ze_3 + \frac{1}{2}(x^2+y^2+z^2)e_\infty + e_0$ \\
\addlinespace
Create sphere $s_1$ centered at $p_2(1,2,3)$ with radius 0.5 (color: red) 
    & $S = C - \frac{1}{2}r^2e_\infty$ \\
\addlinespace
\dots  & \dots \\
\bottomrule
\end{tabular}
\vspace{-20pt}
\end{table}

\begin{table}[h]
\vspace{-20pt}
\caption{GPT-4o Prompt template}
\centering
\label{tab:Prompt template}
\begin{tabular}{c}
\hline
\begin{minipage}{12cm}
You are an expert in GAALOPScript within the geometric algebra domain. Your task is to understand user input and generate GAALOPScript code consisting of three components: optimized computation code, variable assignment statements, and multivectors to be visualized. Below are the task description and GA formula provided by the user: \\
Task Description: \{Task Description\} \\
GA Formula: \{GA Formula\}
\end{minipage} \\
\hline
\end{tabular}
\vspace{-20pt}
\end{table}

\subsection{Compare the task execution capabilities of different LLMs}
This study conducted performance tests on both the baseline model and GA-VisAgent using the same dataset, followed by a detailed analysis of the results. The experimental outcomes reveal that GPT-4o achieves a success rate of only 20\% in generating valid GAALOPScript and ganja visualizations and code. In sharp contrast to this, GA-VisAgent successfully visualized 36 out of 40 formulas, with a success rate of 90\%, significantly higher than that of GPT-4o. The detailed test results are presented in Table~\ref{tab:comparison}.

\begin{table}[htbp]
    \vspace{-20pt}
    \centering
    \caption{Comparison of Models}
    \label{tab:comparison}
    \begin{tabular}{lccc}
        \hline
        \textbf{ID} & \textbf{Total Tasks} & \textbf{Success Tasks} & \textbf{Success Rate (\%)} \\
        \hline
        GA-VisAgent & 40 & 36 & 90 \\
        GPT-4o      & 40 & 8  & 20 \\
        \hline
    \end{tabular}
\end{table}

\subsection{Experiment Case}
In this experimental case, the tested problem is: Visualization formula $S = C - \frac{1}{2} r^2 e_\infty$: In conformal space, create three spheres S1, S2, S3 with centers at X1 (0, 0, 0), X2 (0, 0.4, 0), and X3 (0, 0.45, 0.2) with radii of 0.5, 0.4, and 0.3, respectively, S1, S2, S3 are visualized in blue, red, and green, respectively. Finally, calculate the intersection points x4 and x5 of the three balls and visualize them in yellow. I need Python code.

\begin{figure}[htbp] 
\vspace{-10pt} 
    \centering
    \includegraphics[width=1.0\textwidth]{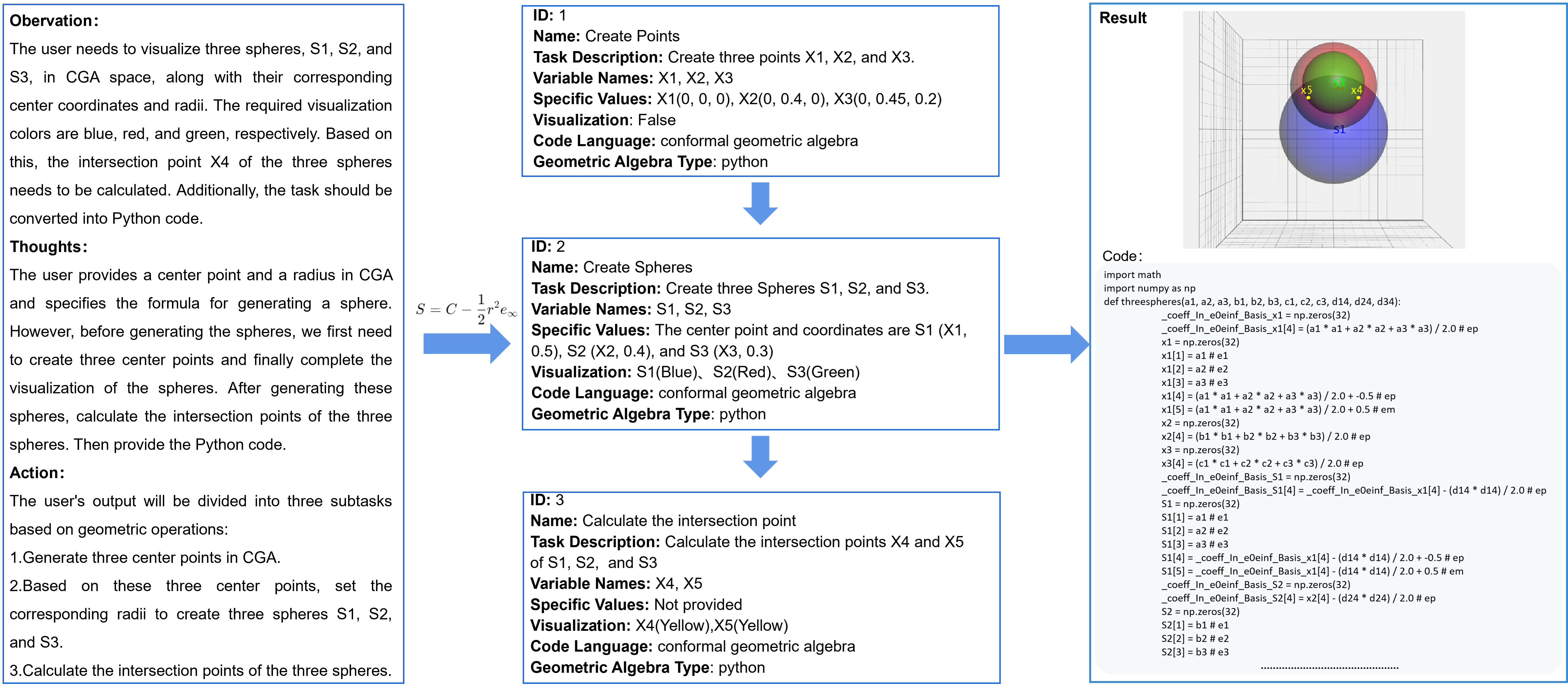}
    \caption{Task decomposition and planning}
    \label{fig:7}
    \vspace{-20pt} 
\end{figure}

Based on the react mechanism, after receiving the GA formulas and GA problems input by the user, the Planner Agent will go through three stages of Observation, Thoughts, and Action to accurately analyze complex problems. In the Observation stage, the planner agent will carefully input GA formulas, identify the variables, operators, and formula structures, and at the same time clarify the specific needs of users, such as visualization, solution, or other operations. Entering the Thoughts stage, the agent thinks about how to disassemble complex tasks into operable subtasks based on the observed information and plans a reasonable execution sequence to ensure that each subtask is logically clear and interconnected. At the Action stage, the agent formally decomposes the task into three ordered subtasks, as shown in Figure~\ref{fig:7}.

After completing the decomposition of subtasks, the Worker Agent will execute them in order, and the visualization code generated by previous subtasks will serve as the context for subsequent subtasks. For example, the content about creating points x1, x2, and x3 in the first subtask will be used as the center point of the sphere in the second subtask, and then combined with the corresponding radius information, the code for spheres S1, S2, and S3 will be generated.




\section{Conclusion and Outlook}
This study proposes GA-VisAgent, a multi-agent framework designed to process problems expressed in natural language and GA formulas, enabling automated formula visualization and code generation. The framework automatically decomposes user-input GA problems into subtasks and efficiently resolves them through agent collaboration. By implementing a subtask classification mechanism and structured output format design, the system accurately handles core GA operations—such as geometric products, rotations, and reflections—significantly reducing users' dual learning burden in both script syntax and GA theory. GAALOPScript serves as the core middleware for code generation and visualization, though the framework's architecture remains generic and can be adapted to other GA software packages with appropriate modifications. Experimental validation demonstrates that the GAALOPScript code generated by GA-VisAgent, after verification and formatting, precisely drives the GAALOP API to successfully execute and visualize complex tasks, including sphere creation and transformation in conformal space.

In the future, the GA-VisAgent framework will commit to further expanding its functional boundaries. Users will no longer need to manually provide complex GA formulas – describing requirements through natural language alone will yield intuitive visualization and code results. This enhancement will significantly lower the barrier to learning and applying GA, enabling non-expert users to effortlessly comprehend and explore abstract GA concepts. Additionally, we will expand GA-VisAgent's compatibility with GA software packages, thus breaking away from its current limitation to GAALOPScript.

\begin{credits}
\subsubsection{\ackname} This work was supported by the National Natural Science Foundation of China (nos. 42230406, 42130103, and 42376223)
\subsubsection{\discintname}
The authors declare no conflict of interest.

\end{credits}
\bibliographystyle{splncs04}
\bibliography{paper432}

@misc{fontijne2007geometric,
  title={Geometric algebra for computer science: An object-oriented approach to geometry (the morgan kaufmann series in computer graphics)},
  author={Fontijne, D and Dorstand, L and Mann, S},
  year={2007},
  publisher={Elsevier}
}

@book{hestenes2012new,
  title={New foundations for classical mechanics},
  author={Hestenes, David},
  volume={15},
  year={2012},
  publisher={Springer Science \& Business Media}
}

@article{angelis2024geometric,
  title={Geometric Algebra Meets Large Language Models: Instruction-Based Transformations of Separate Meshes in 3D, Interactive and Controllable Scenes},
  author={Angelis, Dimitris and Kolyvakis, Prodromos and Kamarianakis, Manos and Papagiannakis, George},
  journal={arXiv preprint arXiv:2408.02275},
  year={2024}
}

@inproceedings{wang2023large,
  title={Large Language Model for Geometric Algebra: A Preliminary Attempt},
  author={Wang, Jian and Wang, Ziqiang and Wang, Han and Luo, Wen and Yuan, Linwang and L{\"u}, Guonian and Yu, Zhaoyuan},
  booktitle={Computer Graphics International Conference},
  pages={237--249},
  year={2023},
  organization={Springer}
}

@article{hadfield2021exploring,
  title={Exploring novel surface representations via an experimental ray-tracer in cga},
  author={Hadfield, Hugo and Achawal, Sushant and Lasenby, Joan and Lasenby, Anthony and Young, Benjamin},
  journal={Advances in Applied Clifford Algebras},
  volume={31},
  pages={1--33},
  year={2021},
  publisher={Springer}
}

@article{li2022ga,
  title={GA-CNN: Convolutional neural network based on geometric algebra for hyperspectral image classification},
  author={Li, Yanping and Wang, Yue and Wang, Rui and Wang, Yi and Wang, Kaili and Wang, Xiangyang and Cao, Wenming and Xiang, Wei},
  journal={IEEE Transactions on Geoscience and Remote Sensing},
  volume={60},
  pages={1--14},
  year={2022},
  publisher={IEEE}
}

@article{thiruvengadam2020time,
  title={Time series, hidden variables and spatio-temporal ordinality networks},
  author={Thiruvengadam, Sudharsan and Tan, Jei Shian and Miller, Karol},
  journal={Advances in Applied Clifford Algebras},
  volume={30},
  number={3},
  pages={37},
  year={2020},
  publisher={Springer}
}

@inproceedings{labunets2004clifford,
  title={Clifford algebras as unified language for image processing and pattern recognition},
  author={Labunets, Valeriy},
  booktitle={Computational Noncommutative Algebra and Applications},
  pages={197--225},
  year={2004},
  organization={Springer}
}

@article{hitzer2022foundations,
  title={Foundations for strip adjustment of airborne laserscanning data with conformal geometric algebra},
  author={Hitzer, Eckhard and Benger, Werner and Niederwieser, Manfred and Baran, Ramona and Steinbacher, Frank},
  journal={Advances in Applied Clifford Algebras},
  volume={32},
  number={1},
  pages={1},
  year={2022},
  publisher={Springer}
}

@article{orouji2019hardware,
  title={A hardware implementation for colour edge detection using prewitt-inspired filters based on geometric algebra},
  author={Orouji, Niloofar and Sadr, Ali},
  journal={Advances in Applied Clifford Algebras},
  volume={29},
  number={2},
  pages={23},
  year={2019},
  publisher={Springer}
}

@article{hildenbrand2010gaalop,
  title={Gaalop—high performance parallel computing based on conformal geometric algebra},
  author={Hildenbrand, Dietmar and Pitt, Joachim and Koch, Andreas},
  journal={Geometric Algebra Computing: in Engineering and Computer Science},
  pages={477--494},
  year={2010},
  publisher={Springer}
}

@article{alves2020efficient,
  title={Efficient development of competitive mathematica solutions based on geometric algebra with gaalopweb},
  author={Alves, Rafael and Hildenbrand, Dietmar and Steinmetz, Christian and Uftring, Patrick},
  journal={Advances in Applied Clifford Algebras},
  volume={30},
  number={4},
  pages={59},
  year={2020},
  publisher={Springer}
}

@article{perwass2009learning,
  title={Learning geometric algebra with clucalc},
  author={Perwass, Christian},
  journal={Geometric Algebra with Applications in Engineering},
  pages={25--48},
  year={2009},
  publisher={Springer}
}

@inproceedings{yao2023react,
  title={React: Synergizing reasoning and acting in language models},
  author={Yao, Shunyu and Zhao, Jeffrey and Yu, Dian and Du, Nan and Shafran, Izhak and Narasimhan, Karthik and Cao, Yuan},
  booktitle={International Conference on Learning Representations (ICLR)},
  year={2023}
}

@article{wang2024survey,
  title={A survey on large language model based autonomous agents},
  author={Wang, Lei and Ma, Chen and Feng, Xueyang and Zhang, Zeyu and Yang, Hao and Zhang, Jingsen and Chen, Zhiyuan and Tang, Jiakai and Chen, Xu and Lin, Yankai and others},
  journal={Frontiers of Computer Science},
  volume={18},
  number={6},
  pages={186345},
  year={2024},
  publisher={Springer}
}

@article{heik2024study,
  title={Study on the application of single-agent and multi-agent reinforcement learning to dynamic scheduling in manufacturing environments with growing complexity: Case study on the synthesis of an industrial IoT Test Bed},
  author={Heik, David and Bahrpeyma, Fouad and Reichelt, Dirk},
  journal={Journal of manufacturing systems},
  volume={77},
  pages={525--557},
  year={2024},
  publisher={Elsevier}
}

@inproceedings{liang2024survey,
  title={A Survey of Multimodel Large Language Models},
  author={Liang, Zijing and Xu, Yanjie and Hong, Yifan and Shang, Penghui and Wang, Qi and Fu, Qiang and Liu, Ke},
  booktitle={Proceedings of the 3rd International Conference on Computer, Artificial Intelligence and Control Engineering},
  pages={405--409},
  year={2024}
}

@article{Wang2025,
  author    = {Wang, Jian and Du, Pei and Zhao, Zhuo and Luo, Wen and Yu, Zhaoyuan and Yuan, Linwang},
  title     = {{GAGPT and Its Application to the Interactive Learning of Geometric Algebra}},
  journal   = {Advances in Applied Clifford Algebras},
  year      = {2025},
  volume    = {35},
  number    = {3},
  pages     = {30},
  doi       = {10.1007/s00006-025-01385-8},
  url       = {https://doi.org/10.1007/s00006-025-01385-8},
  issn      = {1661-4909}
}

@misc{ganja.js,
  doi = {10.5281/ZENODO.3635774},
  url = {https://zenodo.org/record/3635774},
  author = {De Keninck,  Steven},
  title = {ganja.js},
  howpublished = {Zenodo. https://doi.org/10.5281/ZENODO.3635774},
  year = {2020}
}
\end{document}